\documentclass[10pt]{article}
\usepackage[T1]{fontenc}
\usepackage{inputenc}
\usepackage[english]{babel}
\usepackage[nohead,includefoot,margin=2cm]{geometry}
\usepackage[compact,raggedright,small]{titlesec}
\usepackage[affil-it]{authblk}
\usepackage[runin]{abstract}
\usepackage{multicol}
\usepackage{blindtext}
\usepackage{cite}
\usepackage{amsmath}
\usepackage{graphicx}
\usepackage{caption}
\newcommand\given[1][]{\:#1\vert\:}

\addto{\Affilfont}{\small}

\title{\large\bfseries Using Entropy Estimates for DAG-Based Ontologies}
\date{\small (Dated August 2012)}
\author[1]{Andrew S. Warren}
\author[2]{Jo\~ao C. Setubal}
\affil[1]{Virginia Bioinformatics Institute, Virginia Tech, Blacksburg, VA}
\affil[2]{Institute of Chemistry, University of S\~ao Paulo, Brazil}

\begin{document}
  \maketitle

  \begin{abstract}
    \\ \textbf{Motivation:} Entropy measurements on hierarchical structures have been used in methods for information retrieval and natural language modeling. Here we explore its application to semantic similarity.  By finding shared ontology terms, semantic similarity can be established between annotated genes. A common procedure for establishing semantic similarity is to calculate the descriptiveness (information content) of ontology terms and use these values to determine the similarity of annotations. Most often information content is calculated for an ontology term by analyzing its frequency in an annotation corpus. The inherent problems in using these values to model functional similarity motivates our work. 
 \\ \textbf{Summary:} We present a novel calculation for establishing the entropy of a DAG-based ontology, which can be used in an alternative method for establishing the information content of its terms. We also compare our IC metric to two others using semantic and sequence similarity.

  \end{abstract}
  \begin{multicols}{2}
\section{INTRODUCTION}
    The popularity of using ontologies in the analysis of biological data has grown rapidly \cite{jensen_ontologies_2010} since the introduction of the Gene Ontology \cite{ashburner_gene_2000}.  One application of ontology annotations is to determine the semantic similarity between two or more entities, using ontology terms, to convey the extent of functional similarity. For the purposes of this paper, semantic similarity is a measure that quantifies the relatedness of two genes based on their ontology annotations. There are a wide variety of methods for determining semantic similarity based on ontologies \cite{pesquita_semantic_2009}. Many of these methods estimate the information content (IC) of ontology terms. The information content of an ontology term is a measure of how specific and how much information that term conveys. The most widely used method for estimating information content is to calculate suprisal \cite{resnik_using_1995} which, for an ontology term $t$ with probability $p(t)$, is given as \begin{equation}\label{resnik} rIC(t)=-\log p(t)\end{equation}
Here $p(t)$ is usually estimated by the frequency that the term occurs in a set of annotated genes referred to as an annotation corpus. We use $rIC$ to distinguish this from other methods used to calculate information content. This method brings with it some inherent problems. Basing IC, and thus semantic similarity, on a specific corpus runs the risk of biasing on research topics that have been more thoroughly investigated and propagated by electronic annotations. A recent paper on behalf of the Gene Ontology consortium \cite{thomas_use_2012} highlights the drawbacks of making biological inferences using metrics based on a specific annotation corpus. They suggest that such metrics are subject to the ``open world assumption'', i.e.~that our complete biological knowledge is not represented in gene annotations and if a gene lacks a particular GO annotation it does not mean that the gene lacks that function. Similarly, the frequencies of annotations in a particular corpus are subject to the focus of analyses that have been performed on its constituent genes. Depending on the corpus, semantic similarity calculations using term frequency can give little weight to GO terms that convey a very specific function and occur often, or give significant weight to GO terms that are shallow and seldom used. Because gene annotations are in flux, basing the value of terms in a semantic similarity calculation on the current state of annotation makes the resulting values vulnerable to the open world assumption. Semantic similarity measures that use corpus-based probability estimates make sense when comparing or grouping genes relative to the body of knowledge in a corpus, but because they do not directly model the conceptual explicitness of a term, they may lose resolution when it comes to relating how functionally similar genes are.\\
An alternative to estimating the IC of ontology terms from a corpus is using the structure of the ontology itself. Seco et al. \cite{seco_intrinsic_2004} present a method for determining the IC of terms in WordNet taxonomy, a text analysis resource, based on the number of descendants. They argue that the more descendants a concept has the less information it expresses, otherwise there would be no need to further differentiate it. They set the information content of a term $t$ to be
\begin{equation}\label{seco} sIC(t)=1-\frac{\log(\Delta_t+1)}{\log(|N|)} \end{equation} where $|N|$ is the number of terms in WordNet and $\Delta_t$ gives the number of descendants for term $t$. Though this metric relieves any dependency on using a corpus it only values the potential for further refinement without considering the number of ancestor terms that have been specified to define a term. As a result all leaf nodes have the highest possible IC value regardless of their depth in the taxonomy.  This too suffers from a kind of ``open world assumption'' in that leaf terms may be created at any depth within the ontology structure at any time. Intuitively this can lead to overestimation of how informative a term is. A term may be a leaf simply because it implies a subject that has not been extensively developed in the ontology.\\
The issues with existing IC estimation and its use in semantic similarity motivate our method for calculating the entropy of a DAG-based ontology and using it to derive a new information content metric that is independent of individual corpus characteristics.

\section{ENTROPY OF ONTOLOGY}
In this paper, entropy refers to Shannon entropy, which is a measure of uncertainty associated with a message, i.e.~a random variable, given a message source. In this case the message source is an ontology and the message is an annotation. We use ``information content'' to refer to the amount, measured in bits, that an ontology term contributes to the uncertainty that a particular annotation will be made.
Given an ontology $M=\{N,E\}$, let $N$ represent the terms of the ontology, $E$ represent the edges, and $r$ represent the root node. For a term $t$ let $\Delta_t$ represent the descendants of $t$ and $\Pi_t$ represent the ancestors. To estimate the entropy of a DAG-based ontology we calculate the joint entropy of selecting a pair of random terms from the ontology. In other words, we calculate the uncertainty associated with generating a two-term annotation where $X$ and $Y$ are random variables that represent the first and second term respectively.  The entropy of the ontology $M$ is given as \begin{align}\label{entropy} H(M)=H(X,Y_x)=H(X)+H(Y_x \given X)=\nonumber \\-\sum_{x\in X}p(x)\log p(x)-\sum_{x\in X}\sum_{y\in Y_x}p(x)p(y\given x)\log p(y \given x) \end{align}. Where \begin{equation}\label{entropy2} Y_x=\{(M\setminus(\Delta_x\bigcup\Pi_x))\bigcup r\}\end{equation}\\
It is assumed in this case that the ontology given, as with GO, represents a subsumption hierarchy and that any term used in an annotation implies its ancestors. Rather than use the frequency of terms in an annotation corpus, we use a maximum entropy estimate where the probability that the first term is selected for annotation is $p(x)=1/|N|$ and the second term is $p(y \given x)=1/|Y_x|$. For a particular annotation this evaluates the probability of selecting an additional term given that the first term selected precludes its ancestors, descendants, and itself, from being selected as the second term. It is possible to use term frequency in a corpus to estimate the probability as part of this calculation but we do not explore that here. Figure~\ref{dag_entropy} illustrates the entropies assigned to various DAG configurations using this method.

\section{ENTROPY TO INFORMATION}
In information theory the amount of information transmitted by a signal can be computed as the decrease in uncertainty at a receiver and is given as $R=H(x)-H(y \given x)$ where $H(x)$ is the uncertainty as to what signal $x$ will be sent and $H(y \given x)$ is the remaining uncertainty sent after $x$ is received \cite{pierce_introduction_1980}. Although the surprisal of a word is usually taken as its information content, it has been suggested that the reduction in uncertainty after processing a word can be taken as the amount of information it conveys \cite{frank_uncertainty_2010, hale_uncertainty_2006}. We propose an information content metric for DAG-based ontologies that uses the approximate level of uncertainty in the DAG before and after a term is assigned. Given the entropy value for a particular ontology we calculate information content for a term $z$ by estimating the conditional joint entropy supposing that terms previous assignment, then taking the difference in the original and conditional entropies. The information content of the term is given as the decrease in entropy created by excluding $z$ and its ancestors.
 \begin{equation}\label{gic} gIC(z)=\frac{H(X,Y_x)-H(X_z ,Y_{xz} \given z)}{H(X,Y_x)}\end{equation}
 \begin{equation}\label{gic2} X_z = \{(M\setminus\Pi_z ) \bigcup r\}\end{equation}
 \begin{equation}\label{gic3} Y_{xz} = \{(M\setminus(\Delta_x\bigcup\Pi_x\bigcup\Pi_z))\bigcup r\}\end{equation}

\begin{center}
\includegraphics[width=0.9\columnwidth]{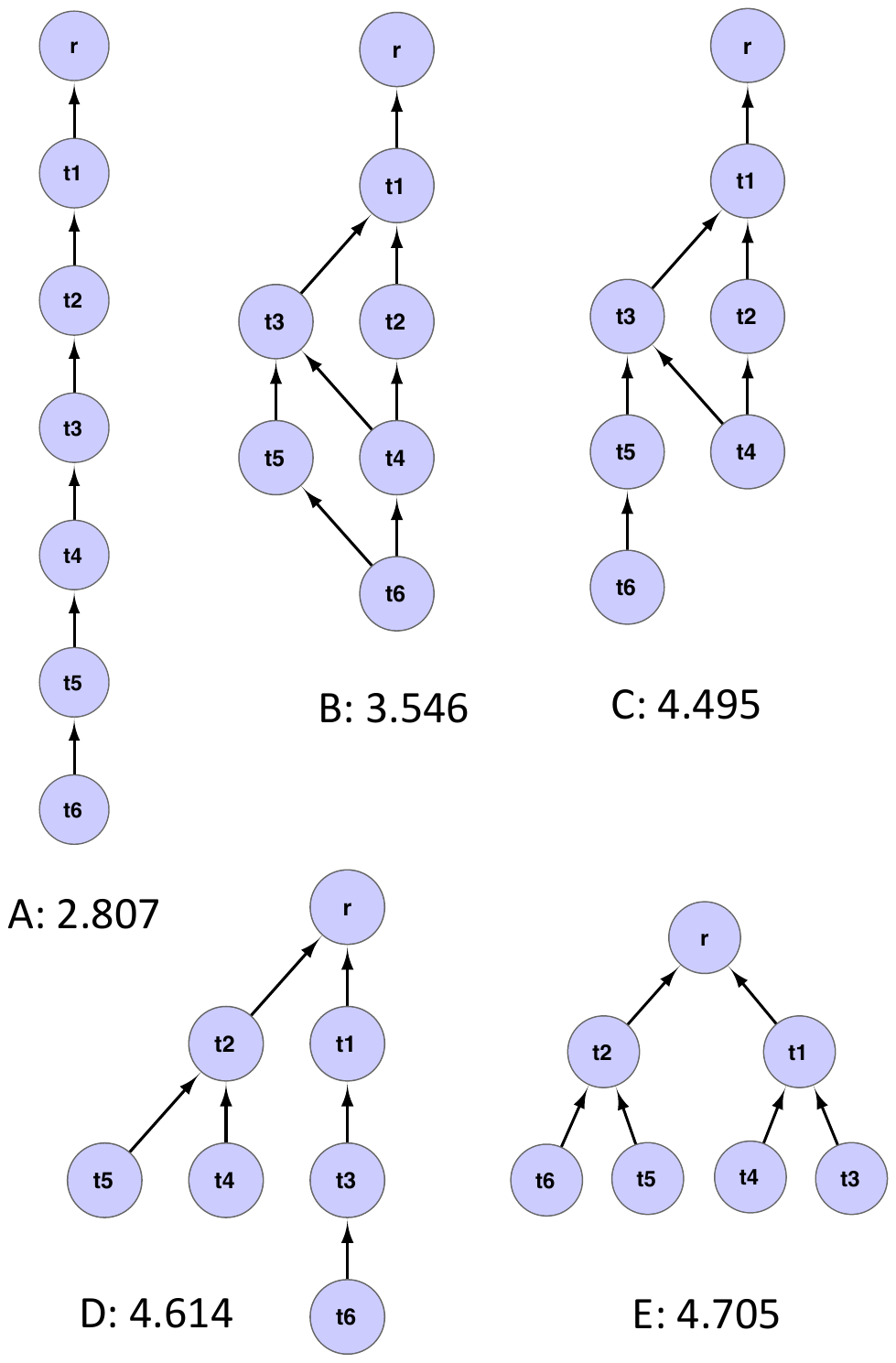}
\captionof{figure}{Entropy values of various DAG configurations.}
\label{dag_entropy}
\end{center}

For any given term our information content measure reflects the information transmitted in its assignment relative to the uncertainty in the ontology. While this metric may be influenced by a particular concept being underdeveloped, it takes into account both the number of ancestors, the number of descendants, and the overall structure of the graph.

\section{SEMANTIC SIMILARITY BENCHMARK}
To determine how well our metric models the biological specificity of ontology terms relative to the other metrics available, we compute and compare semantic similarity values for Gene Ontology annotations using $rIC$, $gIC$, and $sIC$ as input. The performance of semantic similarity is difficult to evaluate because it attempts to quantify the relatedness of concepts that are typically interpreted by humans. As Pesquita et al.~\cite{pesquita_semantic_2009} point out there can be no way to determine the true functional similarity between two gene products. ``If there were, there would be no need to apply semantic similarity in the first place.'' Because of this, methods are usually qualified by examining their behavior relative to a particular application, e.g. clustering, function prediction, cellular location prediction, and protein-protein interaction prediction. \\
To analyze performance we compare the correlation between the average semantic similarities and sequence similarity scores \cite{lord_investigating_2003}. In order to closely tie the resulting semantic similarity value to the value of the information content we use Resnik's maximally informative common ancestor method (SimMax) ~\cite{resnik_using_1995}. To measure sequence similarity we used the relative reciprocal BLAST score (RRBS) \cite{pesquita_metrics_2008}. Given two proteins A and B this is given as 
\begin{equation}\label{rrbs} RRBS = \frac{Bitscore(A,B)+Bitscore(B,A)}{Bitscore(A,A)+Bitscore(B,B)}\end{equation}
To conduct our testing we randomly selected 1,000,000 protein sequences with GO Molecular Function annotations from the UniprotKB proteome. GO annotations were obtained from Uniprot-GOA. Annotations, sequences, and the GO hierarchy were all obtained on July 2011. A BLASTp search was conducted using an all against all approach. Because we wish to measure correlation with sequence similarity and not the effects of shallow annotations, we filtered annotations so that only terms with a minimum edge depth of two or more remained. This resulted in 3,118,974 unique protein pairs with semantic and sequence similarity values. To make the results easier to compare all information content values were normalized using their respective maximum values before determining semantic similarity.
\subsection{Benchmark Analysis}
To account for the variability in annotation, protein sequence, and biological function we calculate average values for semantic and sequence similarity over fixed intervals. We binned protein pairs into groups by sorting according to increasing RRBS, then by increasing semantic similarity value, and finally creating bins at every 1,000 data points. For each bin the mean RRBS and SimMax value was calculated. Figure~\ref{ic_compare} shows the relationship between semantic and sequence similarity using all three IC metrics.

\begin{center}
\includegraphics[width=0.9\columnwidth]{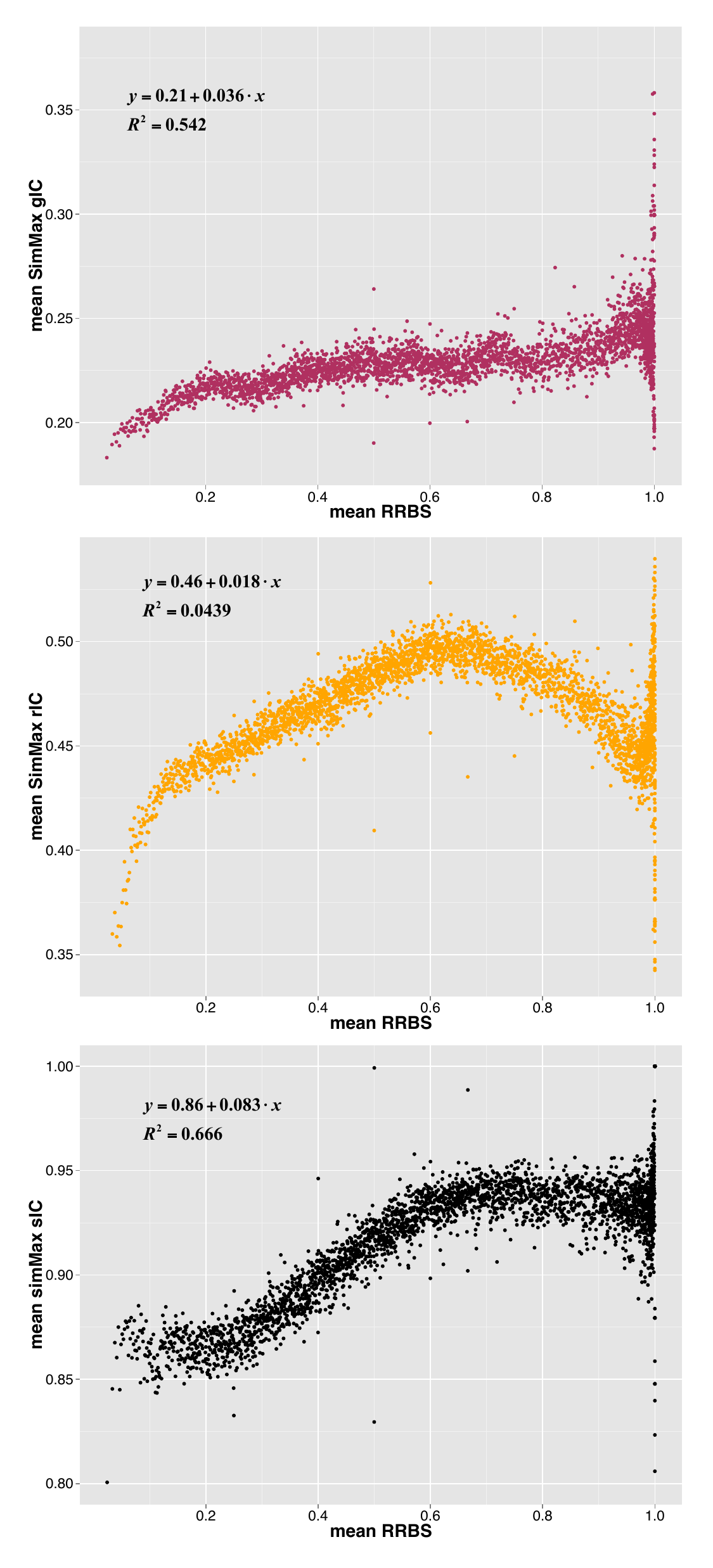}
\captionof{figure}{SimMax gIC, rIC, and sIC values vs. RRBS.}
\label{ic_compare}
\end{center}

In our sample 6\% of the protein pairs (197,391) were identical, i.e.~had an RRBS value of 1. Protein pairs at this level of similarity showed the widest range of SimMax values for all three information content metrics. In Figure~\ref{ic_compare} all three plots are zoomed in on a particular y-value range and exclude some data points with RRBS=1. 
\\As seen in Figure~\ref{ic_compare} only the SimMax value based on $gIC$ shows a general increase across the entire range of sequence similarity. For both $sIC$ and $rIC$, the semantic similarity values peak at an approximate RRBS value of 0.60. In the case of rIC the value substantially decreases after this point. This is caused by shared Molecular Function terms occurring at an increasingly high frequency in the annotation corpus for many of the protein pairs with RRBS on the interval (0.60, 1.0]. Although not tested explicitly here, it stands to reason that pairs of sequences with similarity values above 0.60 are more likely to have their annotations electronically transferred than those below. This can serve as a confounding factor by spreading informative ontology terms to multiple proteins and has been shown to influence semantic similarity results ~\cite{pesquita_metrics_2008}.
\\The SimMax values based on $sIC$ tend to level off above an RRBS value of 0.60. In this case the number of descendants does not provide sufficient resolution to distinguish the shared functions as increasingly specific. 
\\To further characterize the results we also compute the range of semantic similarity values and the $R^2$ for simple linear regression (Table~\ref{r2_table}). The range is calculated as the difference between the maximum and minimum averaged SimMax values taken at the maximum and minimum RRBS values respectively. Some of these values are excluded from Figure~\ref{ic_compare} due to the zoomed nature of the graph. To account for oversampling we excluded all data points with $RRBS=1$ from linear regression. The results based on sIC show a slightly better $R^2$ than those based on $gIC$. However, the range of $sIC$ values is limited which may inhibit resolution of semantic similarity beyond the scale captured by our sequence similarity values.

\begin{center}
\captionof{table}{Benchmark results of SimMax for different IC input}
\begin{tabular}{|c|c|c|c|c|}
\hline
Input & \multicolumn{1}{l|}{Range} & \multicolumn{1}{l|}{Min} & \multicolumn{1}{l|}{Max} & \multicolumn{1}{l|}{$R^2$} \\ \hline
gIC & 0.668 & 0.183 & 0.851 & 0.542 \\ 
rIC & 0.460 & 0.340 & 0.800 & 0.044 \\ 
sIC & 0.199 & 0.801 & 1.0 & 0.666 \\ \hline
\end{tabular}
\label{r2_table}
\end{center}

\section{CONCLUSION}
Determining information content of a term from an annotation corpus conditions on the bias found in both manual and automatic annotations. The variability in annotation quality, method of assignment, and rate of turnover can influence the content of that corpus. We present a method for determining information content independent of annotation trends. This makes our metric more suitable for comparing results across corpora and potentially more reliable in data mining applications because it avoids circularity between the semantic similarity calculation and the annotation corpus being analyzed. In comparison to the corpus-based metric, our information content metric has a higher correlation with sequence similarity and a broader range of values for distinguishing between protein pairs when using Resnik's most informative common ancestor method. This indicates a better representation of functional similarity and shows its potential for enhancing existing analysis methods based on semantic similarity.

\section{ACKNOWLEDGEMENTS}
This project has been funded in whole or in part with Federal funds from the National Institute of Allergy and Infectious Diseases, National Institutes of Health, Department of Health and Human Services, under Contract No. HHSN272200900040C, awarded to BWS Sobral.

\bibliography{dag_entropy_asw}                                                                                                                                                                                                                                                                                                                                                                                                                                                                                                                                                                                                                                                                                                                                                                  

  \end{multicols}

\end{document}